\documentclass{article}
\usepackage{spconf,amsmath,graphicx}

\usepackage[utf8]{inputenc}

\usepackage[table]{xcolor}
\usepackage{times}
\usepackage{epsfig}
\usepackage{rotating}
\usepackage{amssymb}
\usepackage{booktabs}
\usepackage{float}
\usepackage{placeins}
\usepackage{multirow}
\usepackage{balance}
\usepackage{multirow}
\usepackage{subfigure}
\usepackage[flushleft]{threeparttable}
\usepackage{textcomp}
\usepackage{enumitem}

\title{Knowledge Transfer for Melanoma Screening with Deep Learning}

\name{\normalfont{Afonso Menegola}$^\dag$$^\ddag$, Michel Fornaciali$^\dag$$^\ddag$, Ramon Pires$^\circ$,}

\secondlinename{Fl\'{a}via Vasques Bittencourt$^\bullet$, Sandra Avila$^\dag$, Eduardo Valle$^\dag$$^\ast$ 
      \thanks{$^\ddag$Those authors contributed equally.}
      \thanks{$^\ast$Corresponding author: dovalle@dca.fee.unicamp.br.}\vspace{-0.3cm}      
      }

\address{
         $^\dag$RECOD Lab, DCA, FEEC, University of Campinas (Unicamp), Brazil\\
         $^\circ$RECOD Lab, IC, University of Campinas (Unicamp), Brazil\\
         $^\bullet$School of Medicine, Federal University of Minas Gerais (UFMG), Brazil
         }

\begin{document}

\ninept

\maketitle

    \begin{abstract}
        Knowledge transfer impacts the performance of deep learning — the state of the art for image classification tasks, including automated melanoma screening. Deep learning’s greed for large amounts of training data poses a challenge for medical tasks, which we can alleviate by recycling knowledge from models trained on different tasks, in a scheme called \emph{transfer learning}. Although much of the best art on automated melanoma screening employs some form of transfer learning, a systematic evaluation was missing. Here we investigate the presence of transfer, from which task the transfer is sourced, and the application of fine tuning (i.e., retraining of the deep learning model after transfer). We also test the impact of picking deeper (and more expensive) models. Our results favor deeper models, pre-trained over ImageNet, with fine-tuning, reaching an AUC of  80.7\% and 84.5\% for the two skin-lesion datasets evaluated.
    \end{abstract}


    \section{Introduction}

From all skin cancers, melanoma represents just 1\% of cases, but 75\% of deaths\footnote{American Cancer Society: http://www.cancer.org
}. Melanoma's prognosis is good when detected early, but deteriorates fast as the disease progresses. Automated tools may play an important role in providing timely screening, helping doctors focus on patients or lesions at risk, but the disease's characteristics --- rarity, lethality, fast progression, and diagnosis subtlety --- make automated screening for melanoma particularly challenging.

Vast literature exists on automated melanoma screening~\cite{fornaciali2016towards}, but only in the past two years Deep Neural Networks (DNNs)~\cite{krizhevsky2012imagenet} were proposed to improve accuracies~\cite{fornaciali2016towards, deng2009imagenet,codella2015deep, kawaharadeep, masood2015self, premaladha2016novel, sun2016benchmark, demyanov2016classification, cicerodeep, codella2016deep}. DNNs are the state of the art for image classification, but their use for medical images is challenging, since those models require very large training sets (from dozens of thousands, up to several million images)~\cite{shin2016deep}. To bypass that difficulty, most current literature employs \textit{transfer learning}, a technique where a model trained for a given \textit{source task} is partially “recycled” for a new \textit{target task}. Transfer learning ranges from simply using the output of the source DNNs as a feature vector, and training a completely new model (e.g., an SVM) for the target task; until using a pre-trained source DNN to initialize some of the weights of the target DNN, and training the latter as usual~\cite{Yosinski2014}.

\begin{figure}[h]
\begin{center}
    \includegraphics[height=1.5cm]{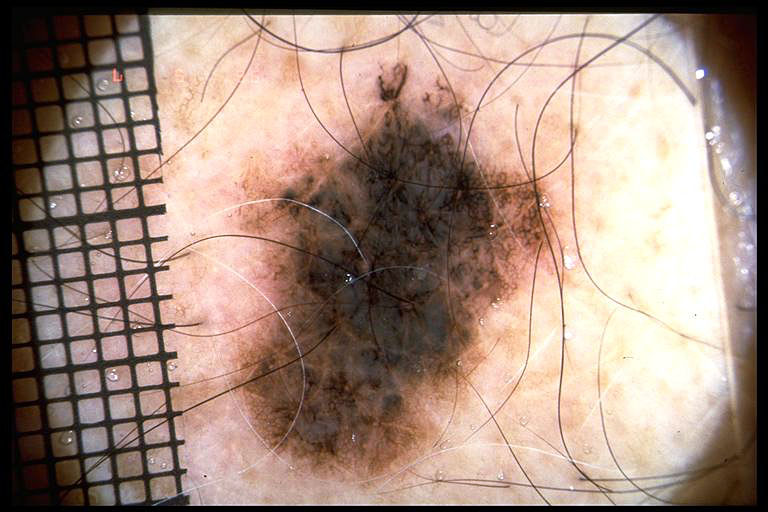}
    \includegraphics[height=1.5cm]{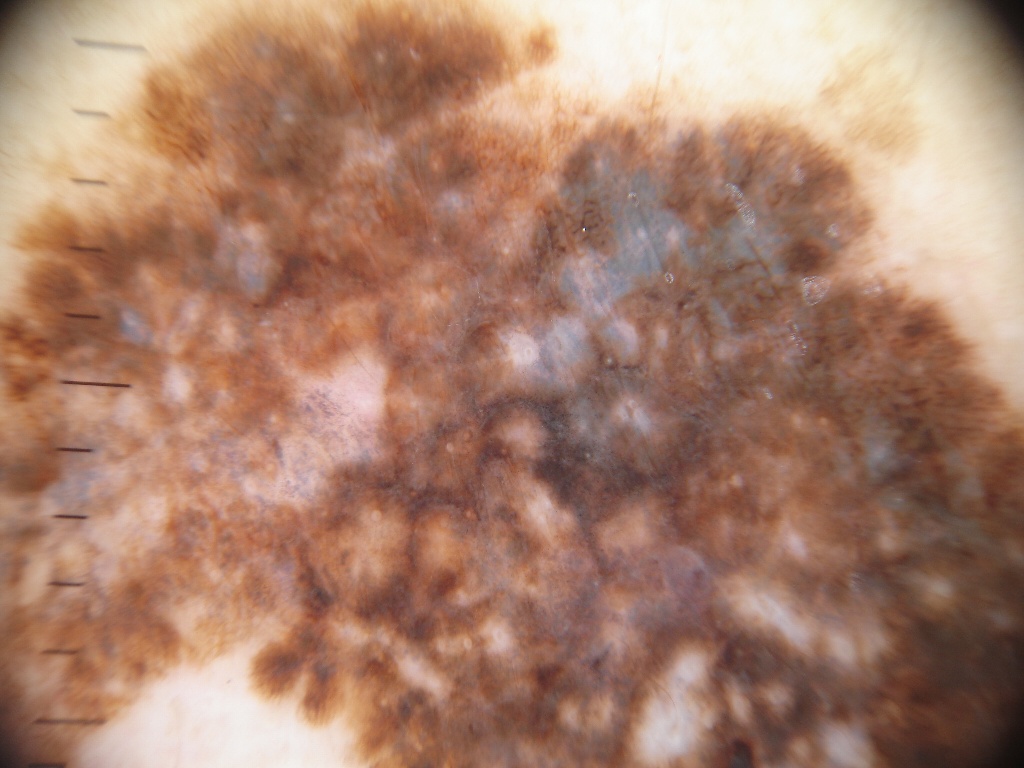}  
    \includegraphics[width=2cm,height=1.5cm]{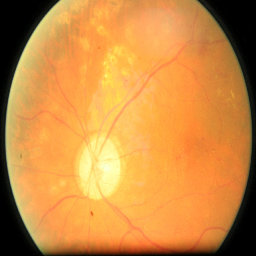}
    \includegraphics[height=1.5cm]{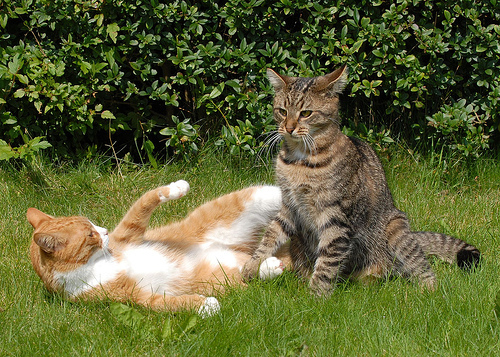}\\ \vspace{-.11cm}
    \subfigure[]{\includegraphics[height=1.5cm]{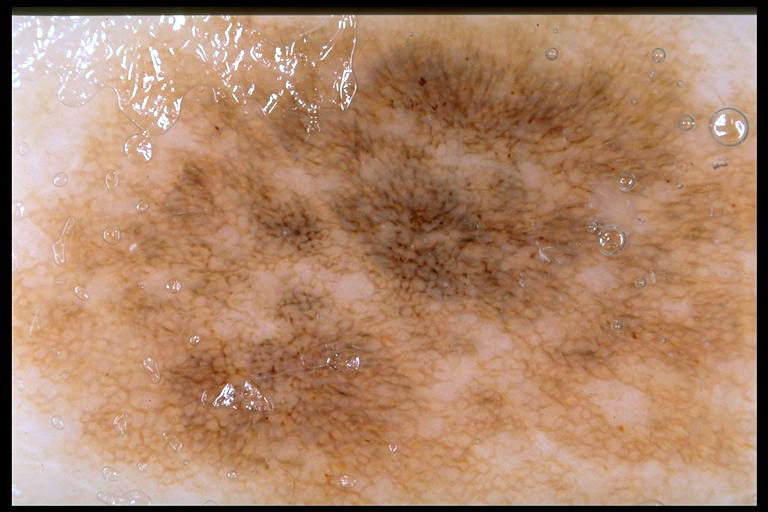}}
    \subfigure[]{\includegraphics[height=1.5cm]{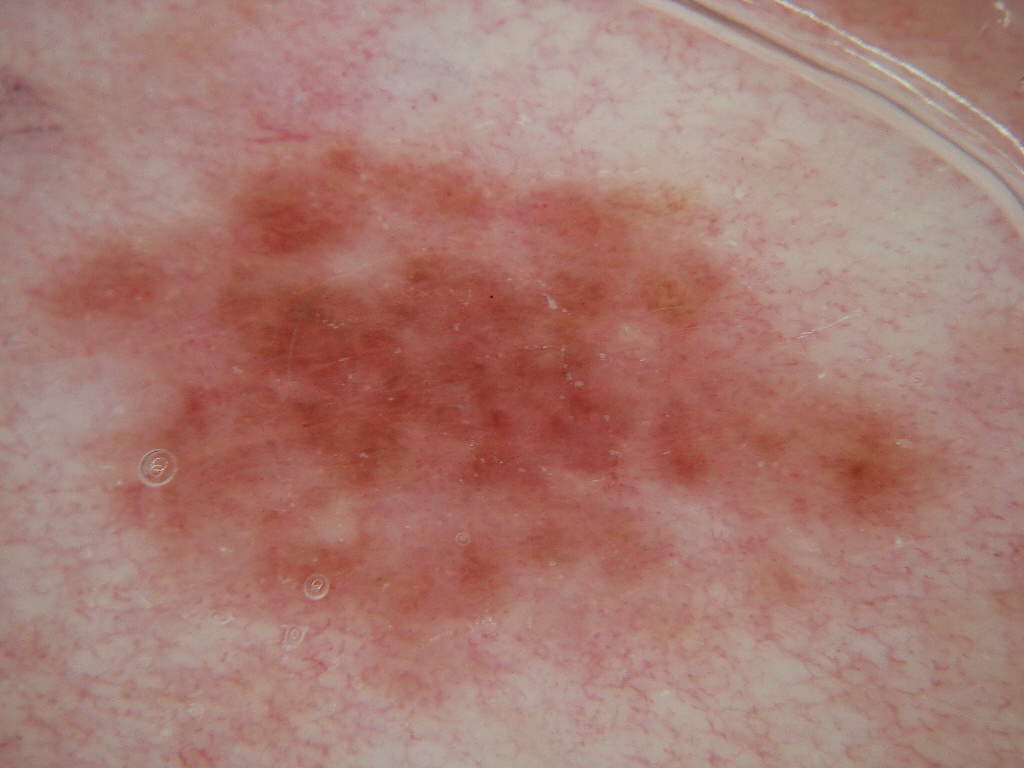}}  
    \subfigure[]{\includegraphics[width=2cm,height=1.5cm]{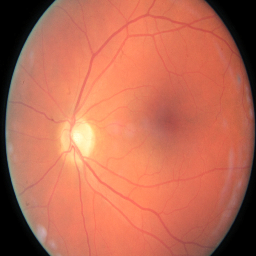}}
    \subfigure[]{\includegraphics[width=2.1cm,height=1.5cm]{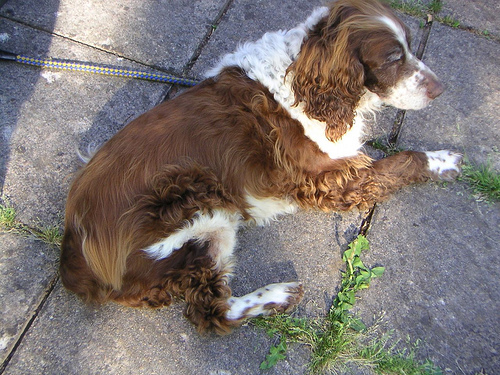}}
	\caption{Samples from datasets used here: (a) Atlas; 
	(b) ISIC; (c) Retinopathy; (d) ImageNet. Each row shows a sample from a different class in the dataset. In this paper, datasets \textit{c} and \textit{d} are source datasets used for transferring knowledge to target models trained in the target task of melanoma screening, trained and evaluated in datasets \textit{a} and \textit{b}.\vspace{-.1cm}}
	\label{fig:datasets}
\end{center}
\end{figure}

Works employing DNNs for melanoma screening either train a network from scratch~\cite{masood2015self, premaladha2016novel, demyanov2016classification}, or transfer knowledge from ImageNet~\cite{codella2015deep, kawaharadeep, sun2016benchmark, cicerodeep, codella2016deep}. The main difference between works is the choice of DNN architecture and implementation framework --- the most common framework is Caffe~\cite{codella2015deep, sun2016benchmark, codella2016deep}, and the most common architectures are  ResNet~\cite{cicerodeep}, DRN-101~\cite{codella2016deep}, AlexNet~\cite{kawaharadeep}, and VGG-16~\cite{sun2016benchmark}. Schemes for artificially augmenting training data, and for transferring learned knowledge also vary.

Segmenting the lesion before feature extraction is a common choice for melanoma screening~\cite{sun2016benchmark, codella2016deep}, although DNNs generally allow forgoing it~\cite{fornaciali2016towards, codella2015deep, kawaharadeep, sun2016benchmark, cicerodeep, fornaciali2014statistical}. In this paper we forgo all preprocessing steps, like lesion segmentation, or artifacts/hair removal. 

Because transfer learning seems so central for successful application of DNNs in melanoma screening, the main contribution of this paper is studying it in more detail. Figure~\ref{fig:datasets} illustrates both the source datasets (Retinopathy and ImageNet) and the target datasets used in our experiments. We attempt to answer the following questions: 
(1) What is the impact of transfer learning? Current art is already convinced it helps, but by how much? 
(2) Is it better transferring from a related (medical) but smaller dataset, from a larger unrelated (general) dataset, or from both? 
(3) Does retraining the transferred network for the new task (known in the DNN literature as \emph{fine tuning}) help, and by how much? Our aim in this paper, thus, is less pushing the envelope on model accuracies, and more increasing our understanding on transfer learning. However, in the spirit of showing that the alternatives evaluated are realistic, we contrast our models to current art and show that they have similar performance. 

Dermoscopic examination is complex, involving many types of lesions, although automated screening is usually advanced as a binary decision, with melanoma as the positive class, and all other lesions as the negative class. However, other skin cancers --- notably basal cell carcinomas --- become a challenge for that model: should they just be put in the negative class with all other lesions? should they be considered together with melanoma in a wholesale “malignant” class? should they be removed altogether from the analysis? or should they be analyzed as a third separate class? A second contribution of this paper is evaluating the impact of such decisions.

    \section{Data and Methods} 
\label{sec:proposed}

A reference implementation with all source code for this paper is available\footnote{https://sites.google.com/site/robustmelanomascreening}. All software libraries mentioned here are linked there.

The datasets employed to train and test the target models (melanoma screening) were the Interactive Atlas of Dermoscopy (Atlas)~\cite{argenziano2002dermoscopy}, and the ISBI Challenge 2016 / ISIC Skin Lesion Analysis towards Melanoma Detection — Part 3: Disease Classification (ISIC)~\cite{gutman2016skin}. Both are available, the former commercially, and the latter free of charge. 

The Interactive Atlas is a multimedia guide (Booklet + CD-ROM) designed for training medical personnel to diagnose skin lesions. The CD-ROM contains 1,000+ clinical cases, each with at least two images of the lesion: close-up clinical image, and dermoscopic image. Most images are 768 pixels wide $\times$ 512 high. Besides the images, each case is composed of clinical data, histopathological results, diagnosis, and level of difficulty. The latter measures how difficult (low, medium and high) the case is considered to diagnose by a trained human. The diagnoses include, besides melanoma (several subtypes), basal cell carcinoma, blue nevus, Clark's nevus, combined nevus, congenital nevus, dermal nevus, dermatofibroma, lentigo, melanosis, recurrent nevus, Reed nevus, seborrheic keratosis, and vascular lesions. There is also a small number of cases classified simply as `miscellaneous'. In this paper we employed only the dermoscopic images, and included all cases except the acral lesions (whose particular appearance poses very specific challenges). Some images contained a black rectangular ``frame'' around the picture, which we cropped automatically with ImageMagick\footnote{http://www.imagemagick.org}. 

The ISBI Challenge 2016 / ISIC Skin Lesion Analysis towards Melanoma Detection — Part 3: Disease Classification~\cite{gutman2016skin} is a subset of 1,279 dermoscopy images from the \textit{International Skin Imaging Collaboration}, an international effort to improve melanoma diagnosis. The Challenge Dataset contains 900 images for training (273 being melanomas) and 379 for testing (115 being melanomas). Ground truth lesion segmentations are available, but we did not use them.

The sources datasets employed for the transfer (pre-training of the DNNs) were the Kaggle Challenge for Diabetic Retinopathy Detection dataset (Retinopathy)\footnote{https://www.kaggle.com/c/diabetic-retinopathy-detection/data}, with a training set of 35,000+ high-resolution retina images taken under varying conditions; and the ImageNet Large Scale Visual Recognition Challenge 2012 dataset (ImageNet), containing ~1M training images labeled into 1,000 categories~\cite{deng2009imagenet}. Both datasets are freely available. Our use of ImageNet dataset was indirect --- because training over it is so time consuming, we opted for employing pre-trained source DNNs.

Figure~\ref{fig:datasets} illustrates those four datasets.

With a single exception, all protocols were based upon the VGG-M model proposed by Chatfield~et~al.~\cite{Chatfield2014}. We also run a single comparison with the VGG-16 model~\cite{simonyan2014very} to evaluate the impact of that deeper (and more expensive) architecture. Because the original implementations are in MATLAB, which was not convenient for our computational infrastructure, we reimplemented those models in Lasagne\footnote{https://lasagne.readthedocs.io/en/latest/} 
and Nolearn\footnote{http://nolearndocs.readthedocs.io/en/latest/lasagne.html}.

In the experiments with transfer learning, we get the source networks pre-trained on ImageNet, train them from scratch on Retinopathy, or fine-tune on Retinopathy the model pre-trained on ImageNet. In the baseline (control) experiment without transfer learning, the networks are trained from scratch. In all networks, we ignore the output layer and employ an SVM classifier to make the decision. We did so for all experiments, including the fine-tuned ones, to avoid introducing extraneous variability.

Whenever training is involved (when we fine-tune or train networks from scratch) we employ the technique of \textit{data augmentation}, which consists in creating randomly modified training samples from the existing ones. Data augmentation is almost always helpful~\cite{krizhevsky2012imagenet, Chatfield2014, sharif2014cnn}, but particularly important when working with the small datasets typical of medical tasks. Augmentation also brings the opportunity to rebalance the classes, alleviating another problem of medical tasks, where the positive class (melanomas) is usually much smaller than the negative (benign lesions). To accomplish both augmentation and balance, we only augment the minority classes (Melanoma, Malignant, and Basal Cell Carcinoma, depending on the experimental design).


The factors investigated are: \textit{transfer learning}, which can be \textit{no transfer} (Atlas from scratch), \textit{from Retinopathy} (Atlas from Retinopathy), \textit{from ImageNet} (Atlas from ImageNet), or \textit{double} (Atlas from Retinopathy from ImageNet); and \textit{fine tuning}, which can be \textit{no FT} (only the final SVM decision layer is trained in the target models), or \textit{with FT} (the target model is retrained for melanoma, before the last layer is stripped and the SVM decision layer is trained).

\begin{table*}
\centering
\caption{Main results (AUC in \%; FT: fine tuning). Surprisingly, transfer from another specific medical task (Retinopathy) is not effective, even if preceded by transfer from a general task (ImageNet). Fine-tuning has major impact and should be considered a necessity. The choice of labeling has a small and somewhat inconsistent impact, that might be due to chance.}
\label{results}
\begin{tabular}{l|cccccc}
\hline
\multirow{3}{*}{Experimental Design} & \multirow{2}{*}{No Transfer} & \multicolumn{2}{c}{From Retinopathy} & \multicolumn{2}{c}{From ImageNet} & Double Transfer\\ 
                                                                                                      &                                    & no FT         & with FT         & no FT   & with FT          &  with FT                                                                                     \\ \hline
Malignant \textit{vs.} Benign                                                                         & 76.0                               & 72.8                    & 76.0                     & 79.1                     & \textbf{82.5}             & 78.8                                                                                  \\ 
Melanoma \textit{vs.} Benign                                                                          & 75.7                               & 73.5                    & 75.3                     & 77.9                     & \textbf{80.9}             & \textbf{80.9}                                                                         \\ 
Melanoma \textit{vs.} Carcinoma \textit{vs.} Benign                                                            & 73.0                               & 71.4                    & 72.8                     & 79.4                     & \textbf{83.6}             & 81.8                                                                                  \\ \hline
\end{tabular}
\end{table*}

The protocol, in detail:
\begin{enumerate}[leftmargin=*]
    \item The weights of the networks are initialized as follows: (a) For experiments without transfer, the weights of the network are initialized as an orthogonal matrix while the biases are initialized as constants ($0.05$); (b) For experiments transferring from ImageNet, the weights are loaded from an available pre-trained network\footnote{http://www.vlfeat.org/matconvnet/pretrained/} (converting from MATLAB format to Nolearn format);  
        (c) For experiments transferring from Retinopathy, the weights are initialized as 1.a, and the network is trained for Retinopathy; (d)  For experiments with double transfer, the weights are loaded from ImageNet as explained in 1.b, and the network is fine-tuned for Retinopathy; 

    \item To reach the input size required by VGG, we resize all images to $224\times224$ pixels, distorting the aspect ratio to fit when needed;
    
    \item We “re-center” the input, as required by VGG, by subtracting the mean of the training dataset. For networks transferred from ImageNet, we used the (precalculated) mean that comes with VGG; for the others, we calculated the mean of the images from the respective training sets; 
    
    \item (Only for the protocols involving DNN training, either from scratch, or with fine-tuning:) We strip the last layer of the source network, add a new softmax layer to make the decision on the skin lesion, and retrain the entire network. When training or fine-tuning the model for Retinopathy, we proceed just like training for Melanoma, except that output softmax units take 5 classes (as specified in Kaggle Competition) instead of 2 or 3 for Melanoma, or 1,000 for ImageNet. We balance the training set by augmenting the data for the minority classes, applying a random transformation (scaling, rotating, or flipping the images). The random number generator is fixed at the very beginning of each experiment, so that the training sequence is the same for all experiments. That makes the comparisons fairer, reducing uncontrolled variability. We train for $60$ epochs with learning rate $10^{-3}$, Nesterov momentum $0.1$, and $\ell_2$-regularization of $5\times10^{-3}$. We perform a 10\% stratified split of the training set for validation. We keep the model that minimizes validation loss; 
    \item We extract the outputs of Group 7/Layer 19, which are vectors of $4,096$ dimensions, and use that output as a feature-vector to describe the images; we $\ell_2$-normalize those vectors, and feed them to a linear SVM classifier (Sklearn implementation) to make the decision on the skin lesion. We separate 10\% of the training set for validation, to choose the SVM regularization parameter, optimizing on the AUC.
\end{enumerate}

We evaluated three experimental designs, varying the labeling of the classes: 
\begin{itemize}
    \item Malignant \textit{vs.} Benign lesions: melanomas and basal cell carcinomas were considered positive cases and all other diagnoses were negative cases; 
    \item Melanoma \textit{vs.} Benign lesions: melanomas were positive cases while all other diagnoses were negative ones, removing basal cell carcinomas;
    \item Basal cell carcinoma \textit{vs.} Melanoma \textit{vs.} Benign lesions: here we have three classes, with all other diagnoses under a single Benign label. 
\end{itemize}

For all designs we employed 5$\times$2-fold cross-validation. Our splits were semi-random, making an effort to balance as much as possible diagnose distributions, to avoid unnecessary variability. 

\begin{table}[t]
\centering
\caption{Results on ISIC, with VGG-16, with transfer learning from ImageNet, with fine tuning. Baselines quoted from the ISIC 2016 competition website (competitors were originally ranked by mAP). The results are provided to show that the models evaluated here are realistic --- in the sense that they are in the same ballpark of performance as current art. All numbers in \%.\vspace{0.05cm}}
\label{results_isic}
\begin{tabular}{l|ccccc}
\hline
 & AUC & mAP & ACC  & SE  & SP  \\ \hline
1\textsuperscript{st} place & 80.4            & 63.7   & 85.5   & 50.7           & 94.1           \\ 
2\textsuperscript{nd} place & 80.2            & 61.9            & 83.1            & 57.3  & 87.2           \\ 
3\textsuperscript{rd} place & 82.6   & 59.8            & 83.4            & 32.0           & 96.1  \\ 
This paper  & 80.7            & 54.9            & 79.2            & 47.6           & 88.1           \\ \hline
\end{tabular}
\end{table}

Our main metric was the Area Under the ROC Curve (AUC); for the design with three classes, we computed three one-vs-one AUCs and reported their average. For the experiment with the ISIC dataset we also report the other measures employed in the competition. We show the results on ISIC for reference purposes, to demonstrate that the models being discussed here are in the same ballpark of performance as the current state of the art (Table~\ref{results_isic}).

    \section{Results}
\label{sec:results}

Our results are in Table~\ref{results}. Fine-tuning improves classification, both when transferring from the small-but-related dataset (Retinopathy), and when transferring from the large-but-unrelated task (ImageNet): that agrees with current literature in DNNs, which almost always endorses fine-tuning. Surprisingly, transfer learning from Retinopathy (also a medical-image classification task) leads to worse results than transferring from the general task of ImageNet, even in combination with the latter. That might indicate that transferring from very specific tasks poses special challenges for overcoming the specialization --- even if the source and target tasks are somewhat related. The best protocol we found was to simply transfer from ImageNet, with fine-tuning. The comparison between DNN architectures shows that --- as usually observed for image classification --- a deeper DNN performs better (Table~\ref{results_architecture}). 

The experimental designs also showed differences in performance: in general it was easier to either group Basal cell carcinomas with Melanomas (Malignant \textit{vs.} Benign), or to consider them as a separate class (Melanoma \textit{vs.} Carcinomas \textit{vs.} Benign), than to ignore them altogether (Melanoma \textit{vs.} Benign). Those results suggest that organizing the labels affects the difficulty of the task, but the explanation for those aggregate numbers might be simply that Basal cell carcinomas are easier to diagnose than Melanomas.
 

We show the results stratified by diagnose difficulty (as indicated by the Atlas itself) in Table~\ref{results_stratified}. Those results show that low-difficult lesions can essentially be solved by current art with relatively high confidence, while for difficult lesions performance is still little better than chance.


\begin{table}[t]
\centering
\caption{Impact of the DNN architecture choice. A deeper model (VGG-16) leads to best results, regardless of the experimental design. All experiments with transfer from ImageNet and fine tuning.}
\label{results_architecture}
\begin{tabular}{l|ccc}
\hline
& \multicolumn{3}{c}{AUC (\%)}                                           \\ 
\multirow{1}{*}{Architecture} & Mal$\times$Ben  & Mela$\times$Ben & Mela$\times$Carc$\times$Ben \\ \hline
VGG-M              & 82.5                     & 80.9                     & 83.6                     \\ 
VGG-16             & \textbf{83.8}            & \textbf{83.5}            & \textbf{84.5}            \\ \hline
\end{tabular}
\end{table}

\begin{table}[t]
\centering
\caption{Results stratified by diagnosis difficulty of test images (Low, Medium or High), for VGG-M, transferring from ImageNet, with fine tuning. All: performance over the whole dataset. Low-, medium-, and high- difficulty cases represent respectively 38.1, 36.3, and 25.6\% of the whole dataset.\vspace{0.05cm}}
\label{results_stratified}
\begin{tabular}{l|cccc}
\hline
& \multicolumn{4}{c}{AUC (\%)} \\
{Experimental Design}  & Low & Medium & High & All \\ \hline
Malignant \textit{vs.} Benign                                                                        & 93.7         & 82.5            & 58.8          & 82.5         \\ 
Melanoma \textit{vs.} Benign                                                                         & 93.0         & 79.6            & 56.6          & 80.9         \\ \hline
\end{tabular}
\end{table}

    \section{Conclusions}
\label{sec:conclusions}

Our results are consistent with current art on DNNs: transfer learning is a good idea, as is fine tuning. Our results also suggest, in line with literature in DNNs, that deeper models lead to better results.  We expected that transfer learning from a related task (in our case, from Retinopathy, another medical classification task) would lead to better results, especially in the \textit{double} transfer scheme, that had access to all information from ImageNet as well. The results showed the opposite, suggesting that adaptation from very specific --- even if related --- tasks poses specific challenges. Still, we believe that further investigation is needed (e.g., can another medical task show better results? can another transfer scheme work?). 

The results suggest that the experimental design is sensitive to the choice of lesions to compose the positive and negative classes, maybe due to the relative difficulty of identifying each of the types of cancer evaluated (Melanomas and Carcinomas).

The results stratified by diagnose difficulty suggest that current art can already deal with the lower and middle spectrum of difficulty, especially considering that human doctors' accuracies might be between 75-84\% \cite{ali2012systematic}. On the other hand, difficult lesions appear \textit{really} hard to diagnose. That suggests a \textit{referability} framework as potentially more fruitful than a \textit{diagnostics} framework. Referring to the doctor both the cases in which the model has high confidence for the positive label, and the \textit{hard} cases (for which the model has low confidence), might be more achievable in the short term than attempting to have high confidence for all cases.

    \section{ACKNOWLEDGEMENTS}
    We are very grateful to the supporters that made this research possible. E. Valle is supported by CNPq/PQ grant 311486/2014-2. M. Fornaciali, R. Pires, and E. Valle are partially funded by Google Research Awards for Latin America 2016. A. Menegola is funded by CNPq;  R. Pires is partially funded by CAPES; S. Avila is funded by PNPD/CAPES. We also acknowledge the support of CNPq (Universal grants 2016); and NVIDIA Corporation, with the donation of a Tesla K40 GPU used for this research. We also thank Prof. Dr. M. Emre Celebi for kindly providing the machine-readable metadata for The Interactive Atlas of Dermoscopy.
    \bibliographystyle{IEEEbib}
    \bibliography{references}

\end{document}